\documentclass[11pt]{phai}

\usepackage{enumitem}
\usepackage{float}
\usepackage{needspace}
\usepackage{placeins}
\usetikzlibrary{fit,backgrounds}

\setlist[itemize]{leftmargin=1.5em,itemsep=0.2em,topsep=0.25em}
\setlist[enumerate]{leftmargin=1.7em,itemsep=0.2em,topsep=0.25em}

\definecolor{rsiBlue}{HTML}{315C8C}
\definecolor{rsiTeal}{HTML}{2A7F7F}
\definecolor{rsiGold}{HTML}{B57A23}
\definecolor{rsiNavy}{HTML}{24415F}
\definecolor{rsiLight}{HTML}{F1F5F8}
\definecolor{rsiChemGreen}{HTML}{118B5E}
\newsavebox{\chemplotbox}

\title{RSI-Anything: Unlocking Recursive\\Self-Improvement with World Models}
\authors{%
{\fontsize{10.2}{13}\selectfont
\mbox{Taoyong Cui\aff{1,2}}\quad
\mbox{Pinxian Zeng\aff{3}}\quad
\mbox{Wanghan Xu\aff{4}}\quad
\mbox{Hao Xiao\aff{5}}\quad
\mbox{Shuhan Xue\aff{1}}\quad
\mbox{Qiang Gao\aff{5}}\quad
\mbox{Wanli Ouyang\aff{2}}\quad
\mbox{Pheng Ann Heng\aff{2,*}}\quad
\mbox{Zhenfei Yin\aff{1,6,*}}\\
\mbox{Yingcheng Wu\aff{1,7,*}}\quad
\mbox{Ling Yang\aff{1,8,*}}%
\par}
\vspace{0.8em}
{\normalsize
\setlength{\fboxsep}{6pt}%
\colorbox{blue!5}{%
  \hspace{8pt}%
  \href{https://github.com/Gen-Verse/RSI-Anything}{%
    \textcolor{blue!65!black}{\textbf{Code: RSI-Anything}}}%
  \hspace{14pt}\textcolor{black!30}{|}\hspace{14pt}%
  \href{https://huggingface.co/datasets/Gen-Verse/RSI-Anything}{%
    \textcolor{blue!65!black}{\textbf{Models: RSI-Anything}}}%
  \hspace{8pt}%
}%
\par}
}
\legend{\textsuperscript{*}Corresponding authors. Author affiliations and correspondence are listed in Appendix~\ref{app:affiliations}.}
\hypersetup{pdfauthor={Taoyong Cui, Pinxian Zeng, Wanghan Xu, Hao Xiao, Shuhan Xue, Qiang Gao, Wanli Ouyang, Pheng Ann Heng, Zhenfei Yin, Yingcheng Wu, Ling Yang}}
\runningtitle{RSI-Anything}

\begin{document}
\maketitle

\begin{abstract}
Recursive self-improvement (RSI) aims to transform interaction experience into persistent improvements, yet external feedback is costly and world-model predictions may become unreliable as policies evolve. We introduce \textbf{RSI-Anything}, a domain-general framework built on JEPA-Anything that uses predictive world models to guide action and learning-data selection. Observed feedback drives policy or workflow improvement and conditional world-model adaptation, while component and compatibility checks govern the retention of updated policy--model pairs.
Across diverse computational and scientific domains, RSI-Anything demonstrates substantial gains. In code repair, with the coder fixed, success increases from $26.3\%$ with a static verifier to $29.2\%$ through predictive selection and $31.2\%$ after world-model adaptation, at comparable test usage. Graphical user interface (GUI) evaluations showed higher task completion, while average actions decrease by $39.4\%$. Across five successive policy versions, laboratory-planning scores more than double from $21.80\%$ to $46.01\%$. In reaction optimization, RSI-Anything achieves $96.15\%$ mean best observed yield versus $82.16\%$ for Bayesian optimization under the same 40-experiment budget. An end-to-end immune-science application further connects workflow evolution and predictive modeling with therapeutic hypothesis generation.
These results demonstrate how predictive world models and retained system updates enable recursive improvement across diverse environments.

\end{abstract}

\begin{figure}[!t]
\centering
\includegraphics[width=\linewidth]{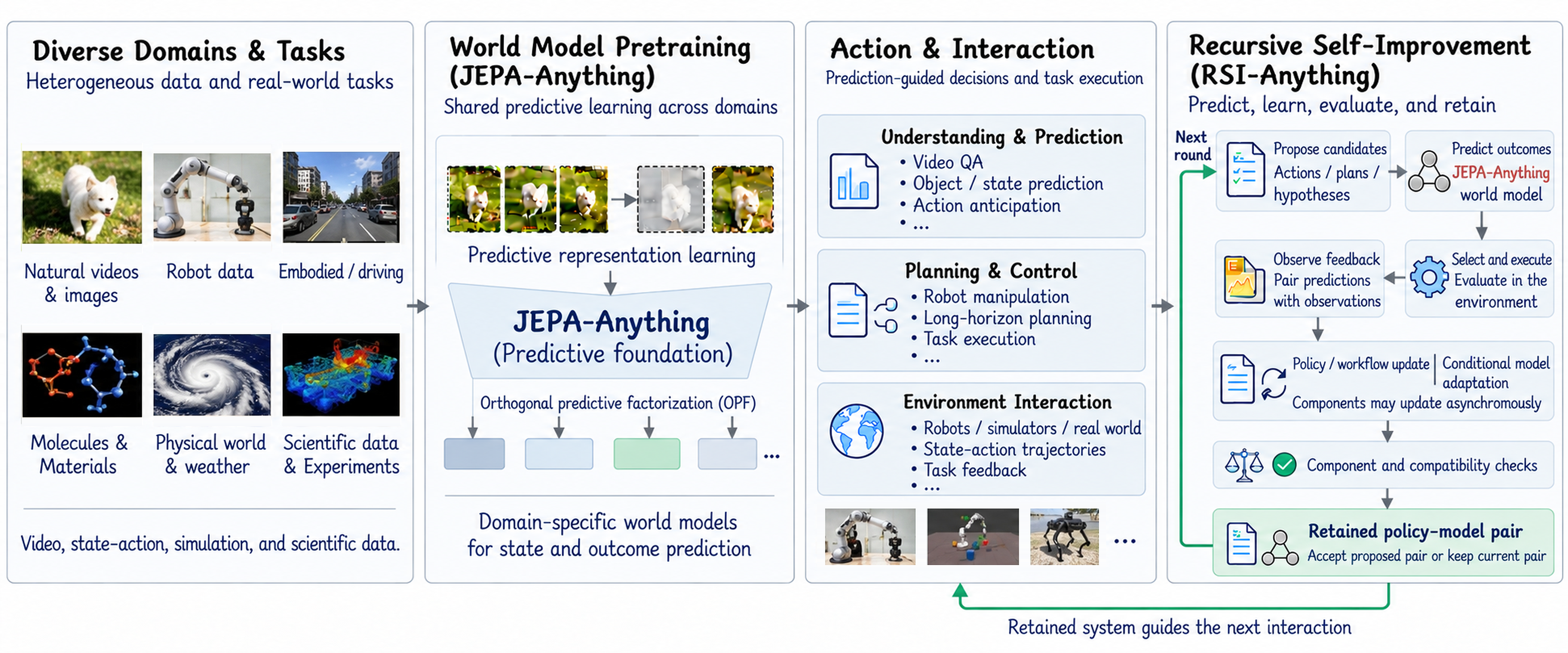}
\caption{\textbf{RSI-Anything overview.} Building on the cross-domain predictive foundation of JEPA-Anything, RSI-Anything uses world-model predictions to support understanding, planning, and action selection across diverse tasks. Interaction feedback drives policy or workflow updates and conditional world-model adaptation. Component and compatibility checks determine which policy--model pair is retained to guide the next round of interaction and learning, closing the recursive self-improvement loop.}
\label{fig:predictive-foundation}
\end{figure}

\section{Introduction}

Recursive self-improvement (RSI) turns interaction experience into lasting improvements in a system's ability to act and learn. Within tasks, feedback refines solutions or guides actions \cite{cobbe2021verifiers,shinn2023reflexion}. Across rounds, self-training and preference optimization support persistent policy or procedural updates \cite{gulcehre2023rest,rafailov2023dpo,zelikman2022star}. Recursiveness requires these updates to be retained and shape subsequent interaction and learning. The central question is how experience from one round improves the system's ability to act and learn in the next.

Learning from external feedback requires executing or evaluating candidates before their consequences can guide improvement, limiting exploration under constrained interaction budgets. World models predict the effects of candidate actions on subsequent states or outcomes \cite{ha2018worldmodels,hafner2025dreamer,schrittwieser2020muzero}, enabling prospective evaluation and planning. Within RSI, these predictions can prioritize feedback collection and guide learning-data selection or construction. World models thus provide a predictive foundation for improvement under limited external feedback.

JEPA-Anything provides a shared framework for world-model construction across heterogeneous domains \cite{jepaanything2026}. Building on joint-embedding predictive architectures \cite{assran2023ijepa,bardes2024vjepa,assran2025vjepa2}, it predicts target representations from observed context. Its orthogonal predictive factorization (OPF) decomposes latent targets into complementary factors, learns dedicated predictors, and recombines their outputs into a complete latent state \cite{jepaanything2026}. This provides a predictive foundation across domains (Figure~\ref{fig:predictive-foundation}). \textbf{However, prediction alone does not establish recursive improvement}: predicted outcomes must guide interaction, while external feedback must produce updates that persist across subsequent rounds.

\textbf{Closing this loop introduces a coupled-update challenge.} Policy updates change the candidate actions and experience distributions on which the world model operates, potentially undermining prediction reliability. Conversely, world-model updates alter candidate rankings, imagined trajectories, and the learning signals guiding policy improvement. Consequently, \textbf{improving either component independently does not guarantee better joint performance}. Reliable RSI therefore requires determining when to update each component, assessing prediction reliability under changing policies, and ensuring that retained policy--model pairs improve subsequent interaction and learning.

\begin{figure}[!p]
\centering
\includegraphics[width=0.99\linewidth,height=0.70\textheight,keepaspectratio]{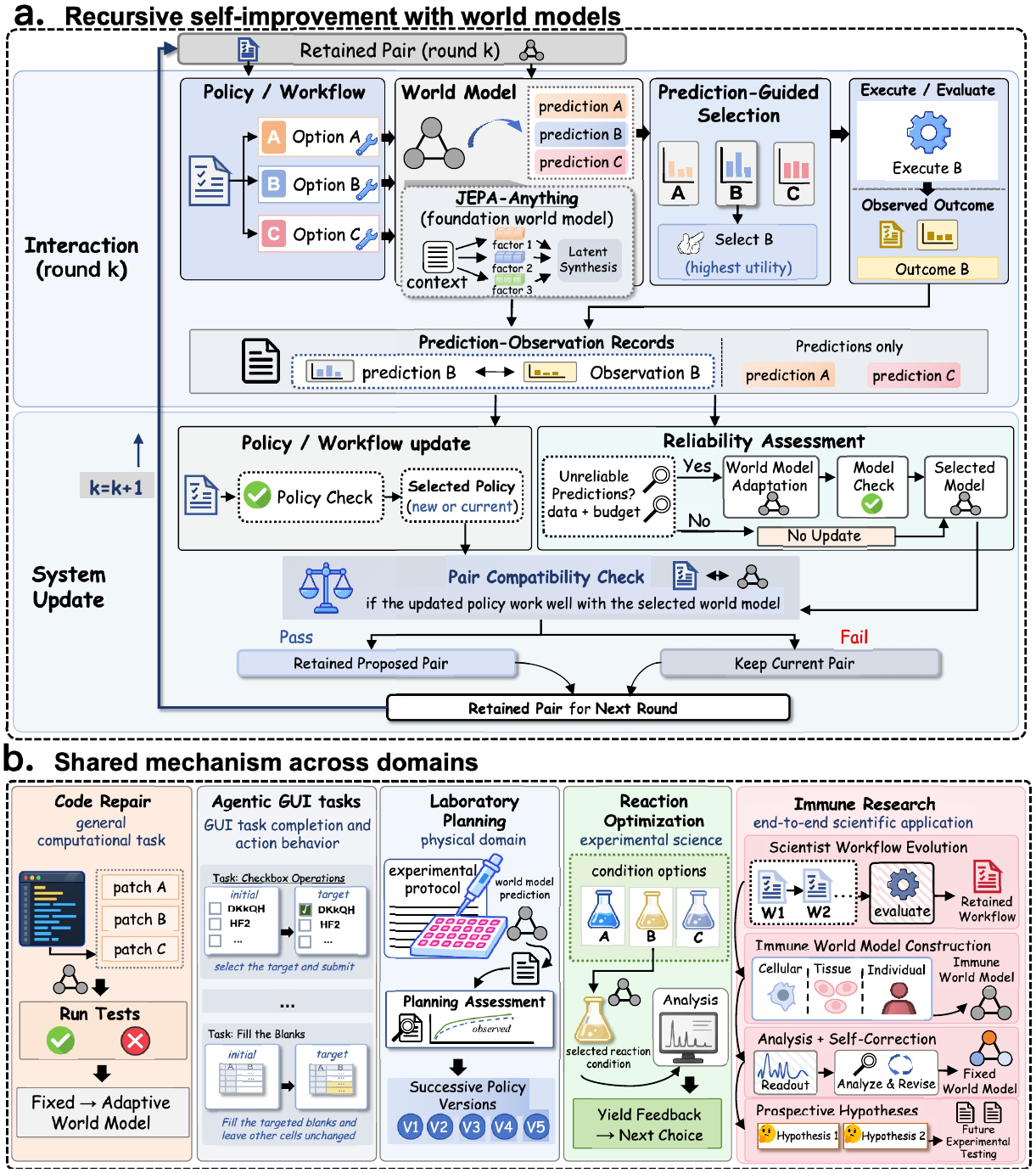}
\caption{\textbf{RSI-Anything couples prediction-guided action with retained system updates.} \textbf{a}, Building on JEPA-Anything, world-model predictions guide candidate selection for external evaluation. Only the executed candidate (B) yields paired prediction--observation feedback. Feedback supports policy or workflow revision and reliability assessment; world-model adaptation depends on prediction reliability, available data, and budget. Component and compatibility checks determine the retained policy--model pair, which shapes candidates and learning experiences in the next round. \textbf{b}, The shared mechanism is instantiated in code repair, GUI interaction, laboratory planning, and reaction optimization. An end-to-end immune-science application connects workflow evolution, immune world-model construction, analysis with self-correction, and prospective hypothesis generation.}
\label{fig:worldmodel}
\end{figure}

To address these challenges, we introduce \textbf{RSI-Anything}, a domain-general framework that integrates predictive world models with feedback-driven recursive improvement (Figure~\ref{fig:worldmodel}). World-model predictions guide action selection, prioritize external evaluations, and support learning-data selection or construction. External feedback drives policy learning or workflow revision and triggers world-model adaptation when reliability checks warrant it. The policy and world model can \textbf{evolve asynchronously}, with component-level checks evaluating individual updates and \textbf{compatibility checks determining which policy--model pair is retained}. This retained pair shapes subsequent candidates and learning experience, closing the recursive improvement loop.

Our evaluations examine whether introducing world-model predictions into RSI improves candidate selection, whether retained policy or model updates yield further improvements, and whether these gains can be achieved under limited external feedback. We investigate these questions across general computational tasks (code repair and GUI interaction), physical-domain planning, and experimental science. We further present an end-to-end application of RSI-Anything in immune science, spanning scientific-agent workflow evolution, immune world-model construction, analysis with self-correction, and prospective therapeutic hypothesis generation \cite{evoiwm2026}.

Our contributions are threefold:
\begin{itemize}
    \item \textbf{World models for recursive self-improvement.} We introduce RSI-Anything, a domain-general framework that integrates predictive world models into recursive self-improvement to anticipate action outcomes and guide experience collection. We formulate persistent improvement at the level of the policy--world-model system, making prediction reliability under evolving policies a central design requirement.
    \item \textbf{A recursive update-and-retention mechanism.} We connect prediction-guided action and learning-data selection with feedback-driven policy learning or workflow revision and conditional world-model adaptation. Component and compatibility checks determine which policy--model pair is retained, allowing asynchronous updates. The retained pair shapes subsequent candidates and learning experience, linking each accepted system revision to the next round of improvement.
    \item \textbf{Empirical evidence across domains.} With the coder fixed, we demonstrate successive gains from introducing world-model-guided selection and adapting the predictor at similar online test usage. Further evaluations show higher GUI task completion, improvement across five successive laboratory-planning policy versions, and higher reaction yields than Bayesian optimization at matched experiment counts. An end-to-end immune-science application illustrates how RSI-Anything connects retained workflows and predictive models with evidence-guided analysis and hypothesis generation.
\end{itemize}

\section{RSI-Anything Framework}

RSI-Anything uses world-model predictions to guide action selection and feedback collection, then uses the resulting experience to propose and retain system updates (Figure~\ref{fig:worldmodel}). The retained policy--model pair determines subsequent candidates and the experience available for further improvement. The shared mechanism connects prediction-guided selection, feedback-driven updates, and component and pair acceptance. Domain implementations specify the models, action encodings, outcome readouts, and enabled pathways, including uncertainty-aware selection, multistep planning, and imagined-trajectory learning. The full update-and-retention protocol is defined below; Table~\ref{tab:implementation} identifies the update scope reported for each application.

\subsection{Problem formulation}

For domain $d$, the task specification is

\begin{equation}
\mathcal{T}_d=(\mathcal{X}_d,\mathcal{S}_d,\mathcal{A}_d,\mathcal{W}_d,\mathcal{V}_d,\mathcal{M}_d,\mathcal{C}_d,\mathcal{B}_d),
\end{equation}

where $\mathcal{X}_d$ describes observations and context, $\mathcal{S}_d$ the task state, $\mathcal{A}_d$ available actions or interventions, $\mathcal{W}_d$ the world-model adapter, and $\mathcal{V}_d$ the verifier for candidate validity and predicted utility. The remaining terms specify evaluation metrics, operational and safety constraints, and budgets for computation, simulation, and external interaction. Executing or evaluating a selected action yields observed feedback. The adapter links this feedback to the stored prediction, the action and context that produced it, and the evidence source.

We use $t$ for action steps within a task and $k$ for update-and-acceptance cycles. At cycle $k$, the accepted system is $(\theta_k,\phi_{d,k})$. The policy configuration $\theta_k$ determines how candidates are generated or selected and can comprise trainable parameters or a persistent agent workflow; $\phi_{d,k}$ specifies the current predictor. A cycle can aggregate feedback from several action steps, trajectories, or tasks. It ends after update decisions and acceptance checks, even if both components retain their current versions. Thus, $k$ counts update cycles rather than successful revisions.

Within a run, the task objective, state and action definitions, interfaces, scoring criteria, constraints, acceptance and stopping rules, and total budget limits remain fixed. Remaining budgets decrease with resource use. Task states, candidate pools, feedback records, and accepted component versions can evolve. An action or observation changes the task state or available information; an accepted change to $\theta_k$ or $\phi_{d,k}$ changes the system used in subsequent decisions. Extending the feedback history alone is not a policy or model revision. Table~\ref{tab:update-scope} summarizes this distinction.

\begin{table}[H]
\centering
\footnotesize
\setlength{\tabcolsep}{4pt}
\caption{Fixed settings, update rules, and retained state in a recursive run.}
\label{tab:update-scope}
\begin{tabular}{>{\raggedright\arraybackslash}p{0.18\linewidth}
                >{\raggedright\arraybackslash}p{0.39\linewidth}
                >{\raggedright\arraybackslash}p{0.35\linewidth}}
\toprule
\textbf{Object} & \textbf{Change rule} & \textbf{Carried forward} \\
\midrule
Task and evaluation specification & Fixed for the run & Task objectives and definitions, interfaces, evaluation criteria, constraints, budget limits, and acceptance and stopping rules \\
Task state and experience & Actions and observations revise the state and extend prediction--observation records & Current task state for subsequent actions; accumulated records for selection and learning \\
Policy configuration $\theta$ & Feedback-driven policy learning or workflow revision proposes a candidate version; component and pair checks govern acceptance & Accepted policy configuration; otherwise the current version \\
World model $\phi_d$ & Prediction unreliability, sufficient data and budget trigger an adaptation attempt; component and pair checks govern acceptance & Accepted predictor; otherwise the current version \\
\bottomrule
\end{tabular}
\end{table}

\subsection{Predictive foundation}

JEPA-Anything supplies a context--target predictive design in which orthogonal predictive factorization (OPF) organizes latent targets into complementary factors, predicts them from context, and synthesizes a complete latent state \cite{jepaanything2026}. Its learning formulation combines factor prediction with orthogonality and representation regularization. This is the inherited model-construction foundation. Subsequent RSI adaptation specifies its own objectives and trainable parameter subset for the instantiated predictor.

At cycle $k$, encoding and action-conditioned latent prediction are written as

\begin{align}
z_t &= E_{\phi_{d,k}}(s_t,c_d),\\
\widehat z_{t+1} &= F_{\phi_{d,k}}(z_t,a_t,c_d).
\end{align}

Here $s_t$ is the task state, $a_t$ a candidate action, and $c_d$ the domain context. The latent predictor $F$ represents factor prediction and synthesis when OPF is instantiated. The readout $R$ converts the predicted latent target $\widehat z_{t+1}$ into an outcome prediction for selection and feedback comparison. These symbols denote functional roles; adapters specify the action encoding and can incorporate it into the predictor input or context.

A deterministic readout returns a point prediction, whereas a probabilistic readout returns a distribution. These are alternative domain instantiations of the outcome interface:

\begin{align}
\widehat o_{t+1} &= R_{\phi_{d,k}}(\widehat z_{t+1},c_d)
\quad\text{(point readout)},\\
p_{\phi_{d,k}}(o_{t+1}\mid s_t,a_t,c_d)
&= R_{\phi_{d,k}}(\widehat z_{t+1},c_d)
\quad\text{(probabilistic readout)}.
\end{align}

The domain specification states how the readout is converted into task utility and compared with observed feedback. Point predictions use the corresponding outcome error; probabilistic predictions can additionally use distribution-sensitive scores or calibration measures. Any uncertainty estimator used in selection or training admission is specified separately. Reliability concerns agreement with observed outcomes and the coverage of those observations.

One-step selection requires only an outcome readout. Multistep planning additionally requires an adapter that propagates a planning state. When this transition interface is available, a rollout of horizon $H$ under goal $g$ has the form

\begin{equation}
p_{\theta_k,\phi_{d,k}}(\widehat\tau\mid s_0,g,c_d)
=\prod_{t=0}^{H-1}\pi_{\theta_k}(a_t\mid s_t,g)\,
p_{\phi_{d,k}}(s_{t+1},o_{t+1}\mid s_t,a_t,c_d).
\end{equation}

The propagated $s_{t+1}$ remains a predicted planning state until externally checked. The joint transition readout is needed only for applications that perform multistep rollouts.

\subsection{Prediction-guided selection}

At state $s_t$ and goal $g$, the current policy proposes $\mathcal K_t=\operatorname{Propose}_{\theta_k}(s_t,g)$. Each candidate receives an outcome prediction or, when supported, a bounded rollout $\widehat\tau$. The verifier records validity $v_\tau$, utility components $\mathbf r_\tau$, evidence provenance, and any available model or verifier uncertainty. Validity and operational constraints determine which candidates may be executed; the selection rule determines which feasible candidates receive feedback.

For utility-based ranking, the candidate score is

\begin{equation}
R_d(\widehat\tau)=\sum_j w_{d,j}r_{\tau,j}-\Delta_d(\widehat\tau),
\end{equation}

where $\Delta_d(\widehat\tau)=\lambda_d U(\mathbf u_\tau)$ when uncertainty penalization is enabled and $\Delta_d(\widehat\tau)=0$ otherwise. Invalid candidates are excluded before ranking. The domain configuration specifies utility weights, the implemented ranking or acquisition rule, batch size, and estimated feedback costs. Selection returns a feasible candidate or batch whose evaluation cost fits the remaining feedback budget. In code repair, the instantiated rule aggregates predicted execution-class probabilities to rank four proposed revisions before one revision is executed.

An explicitly enabled information-seeking acquisition rule can instead prioritize uncertain candidates for evaluation. Its definition and use are reported separately from utility-based ranking. Once feedback is obtained, the observed outcome is stored separately from the prediction used to select the candidate.

\paragraph{Evidence and imagined training data.}

RSI-Anything distinguishes deterministic rule checks (E0), learned-model predictions and rollouts (E1), specialized simulators or higher-fidelity calculations (E2), and external measurements or experiments (E3). Predicted and observed fields retain their own provenance. Evidence category records origin and intended fidelity; sufficiency for training is determined by the admission criteria below.

For applications using imagined trajectories for learning, admission is governed separately from external-evaluation selection:

\begin{align}
e_\tau &= \mathbf 1\{v_\tau=1\}
\mathbf 1\{R_d(\widehat\tau)\ge r_d^{\min}\}
C_d^U(\widehat\tau)
\mathbf 1\{\ell_\tau\in\mathcal E_d^{\mathrm{train}}\},\nonumber\\
\mathcal D_{\mathrm{sim},k} &= \operatorname{Curate}\{\widehat\tau:e_\tau=1\}.
\end{align}

Here $\ell_\tau$ records the evidence used to support training admission. The uncertainty check $C_d^U$ tests $U(\mathbf u_\tau)\le u_d^{\max}$ when enabled and returns one when omitted. The permitted evidence sources and thresholds are specified for the application. If imagined learning is disabled, $\mathcal D_{\mathrm{sim},k}$ is empty. This training filter does not prevent external evaluation of uncertain candidates.

\subsection{Feedback-driven improvement}

External feedback extends history $\mathcal H_k$ with prediction--observation pairs linked to the generating system version. Records selected for policy improvement form $\mathcal D_{\mathrm{policy},k}$; records selected for predictor adaptation form $\mathcal D_{\mathrm{adapt},k}$. The two sets serve different objectives and need not contain the same records. Policy improvement changes how candidates are generated or selected; predictor adaptation changes how their consequences are forecast.

\paragraph{Policy learning or workflow revision.}

Feedback must be converted into a signal appropriate to the update object. Demonstration-based learning uses selected actions or trajectories as targets; preference learning compares alternatives using their evaluated outcomes; workflow search uses evaluation feedback to rank persistent agent-workflow configurations. Each application selects the update route appropriate to its policy representation and feedback. Eligible imagined trajectories can supplement policy data when that route is enabled. The policy update seeks better task behavior under the declared objective and constraints.

Let $u_k^\pi$ indicate an enabled policy-update attempt at its declared schedule or feedback trigger, with the required learning evidence and sufficient budget for the proposed attempt. A fixed-policy setting has $u_k^\pi=0$:

\begin{equation}
\theta'_{k+1}=
\begin{cases}
\operatorname{Update}(\theta_k,\mathcal D_{\mathrm{policy},k},\mathcal D_{\mathrm{sim},k}), & u_k^\pi=1,\\
\theta_k, & u_k^\pi=0.
\end{cases}
\end{equation}

The instantiated route specifies how feedback constructs training targets or workflow scores, which parameters or configuration fields can change, and when updates are attempted.

\paragraph{Conditional world-model adaptation.}

Adaptation uses observed targets to improve the corresponding predictions. Outcome labels can supervise an outcome readout; latent prediction targets require an observation that the domain target encoder can represent. A scalar outcome supplies an outcome target rather than a fully observed target state, although its loss can update upstream parameters when they are trainable. The adaptation specification identifies the available target, its predictive loss, and the trainable parameter subset. These choices can differ from those used in initial model construction.

The reliability protocol requires predictions attributable to the current model version. For each historical observation included in the assessment, it uses either a stored prediction generated by that version or a new prediction obtained by applying the current model to the recorded input. Earlier-version predictions remain available for analysis of past decisions but are not pooled directly into the current error estimate. The implementation specifies record selection, version handling, and whether evaluation records have also been used for training.

Reliability is assessed from feedback history $\mathcal H_k$ using this version rule and the candidate range considered in cycle $k$, denoted $\mathcal K_k$. Let $B_{\mathrm{update},k}$ denote the remaining budget available for the proposed adaptation and $\widehat c_{\mathrm{adapt},k}$ its estimated cost in the same units. For an enabled adaptation pathway, the assessment and trigger are

\begin{align}
\rho_k^W &= \operatorname{Reliability}(W_{\phi_{d,k}};\mathcal H_k,\mathcal K_k)
\in\{\mathrm{reliable},\mathrm{unreliable},\mathrm{insufficient}\},\\
u_k^W &= \mathbf 1\{\rho_k^W=\mathrm{unreliable}\}
\mathbf 1\{\mathcal D_{\mathrm{adapt},k}\ \mathrm{is\ sufficient}\}
\mathbf 1\{B_{\mathrm{update},k}\ge\widehat c_{\mathrm{adapt},k}\}.
\end{align}

The assessment applies the specified error, calibration, drift, and coverage criteria supported by the implementation. Because feedback comes from selected candidates, its conclusions are limited to the observed and covered candidate range. Insufficient coverage yields an insufficient assessment rather than evidence of reliability and does not activate the adaptation trigger. Subsequent selection follows the declared evaluation rule and remaining feedback budget.

\begin{equation}
\phi'_{d,k+1}=
\begin{cases}
\operatorname{Adapt}(\phi_{d,k},\mathcal D_{\mathrm{adapt},k}), & u_k^W=1,\\
\phi_{d,k}, & u_k^W=0.
\end{cases}
\end{equation}

This trigger governs adaptation after initial model construction; a disabled pathway sets $u_k^W=0$. The budget specification identifies the constrained resource and accounts for adaptation, historical re-evaluation, and acceptance checks. When policy and model updates share a budget, each attempt uses the amount remaining after prior allocations. External feedback use is reported separately from update and evaluation computation. Either component can remain unchanged while the other is updated; proposed candidates enter the retained system only after acceptance.

\subsection{Recursive retention}

The full retention protocol compares proposed components and their assembled pair with the current versions under predeclared acceptance criteria. Component checks address policy task utility and constraints, or predictor quality on relevant targets. The pair check addresses the dependency between them: a policy can generate candidates for which an otherwise improved predictor is unreliable, and a predictor can change which candidates are selected. Pair acceptance therefore concerns the decision behavior of the assembled system and predictor suitability on candidates proposed by the selected policy, beyond isolated component quality.

For compactness, define the component decisions as

\begin{align}
G_k^\pi &= Q_\pi(\theta'_{k+1};\theta_k,\mathcal D^\pi_{\mathrm{gate},k}),\\
G_k^W &= Q_W(\phi'_{d,k+1};\phi_{d,k},\mathcal D^W_{\mathrm{gate},k}).
\end{align}

The arguments explicitly identify the current version as the comparator. Each check returns one only when its specified criteria are met. The policy and model versions passed to pair evaluation are

\begin{align}
\bar\theta_{k+1}&=
\begin{cases}
\theta'_{k+1}, & u_k^\pi=1\ \land\ G_k^\pi=1,\\
\theta_k, & \text{otherwise},
\end{cases}\\
\bar\phi_{d,k+1}&=
\begin{cases}
\phi'_{d,k+1}, & u_k^W=1\ \land\ G_k^W=1,\\
\phi_{d,k}, & \text{otherwise}.
\end{cases}
\end{align}

When at least one component changes, the pair decision compares the assembled candidate with the current pair under the same prescribed evaluation conditions; otherwise $G_k^{\mathrm{pair}}=1$ without a new evaluation:

\begin{equation}
G_k^{\mathrm{pair}}=Q_{\mathrm{pair}}\!\left(
(\bar\theta_{k+1},\bar\phi_{d,k+1});
(\theta_k,\phi_{d,k}),\mathcal D_{\mathrm{compat},k}\right).
\end{equation}
\begin{equation}
(\theta_{k+1},\phi_{d,k+1})=
\begin{cases}
(\bar\theta_{k+1},\bar\phi_{d,k+1}), & G_k^{\mathrm{pair}}=1,\\
(\theta_k,\phi_{d,k}), & \text{otherwise}.
\end{cases}
\end{equation}

The evaluation specification fixes the task metrics, prediction checks, constraints, feedback allowance, and acceptance thresholds. Corresponding task conditions can be matched even when the two policies generate different candidates. Failure of a component check retains its current version; failure of the pair check retains the entire current pair. If neither component changes, the current pair carries forward without a new pair comparison. Updates are retained when the candidate pair meets the prespecified acceptance criteria under matched evaluation conditions.

Update data construct candidate versions, gate data select which versions are retained, and any designated held-out audit set is reserved for final reporting.

The retained pair and accumulated feedback history enter the next cycle. A changed policy can alter which candidates are proposed; a changed predictor can alter rankings and learning signals. Together, these changes affect which actions receive feedback and what experience is available for subsequent updates. This dependence of later interaction and learning on retained revisions defines the recursive mechanism. The six operations in Figure~\ref{fig:worldmodel} therefore connect candidate proposal and prediction to external selection, observation, update construction, and system retention.

\Needspace{12\baselineskip}

\section{Results}

We examined how world-model predictions improve action selection, how retained updates support further improvement, and how prediction and feedback guide decisions under limited interaction budgets. Code repair and GUI interaction cover general computational tasks, alongside physical-domain laboratory planning and experimental reaction optimization. Within these settings, we evaluate predictive selection, model adaptation, task completion, successive policy versions, and search at matched experiment counts. We then examine an end-to-end application of RSI-Anything in immune science, connecting scientific-agent workflow evolution, world-model construction, and scientific analysis.

\subsection{World-model-guided selection and adaptation improve code repair}

We evaluated two contributions of RSI-Anything in code repair: introducing world-model predictions for candidate selection and obtaining further gains by updating the predictor with execution feedback. The ReasonFlux-Coder-7B policy was held fixed \cite{wang2025cure}. We compared RSI-Anything with a fixed world model (CURE-WM-Lite) and RSI-Anything with world-model adaptation. All selection methods used the same coder and candidate budget: four revisions were proposed before one was selected for execution. Evaluation covered 200 CodeContests tasks \cite{li2022alphacode} across three seeds and up to three feedback rounds, with online tests supplying feedback and hidden tests measuring final success.

\begin{figure}[!htbp]
\centering
\includegraphics[width=\linewidth]{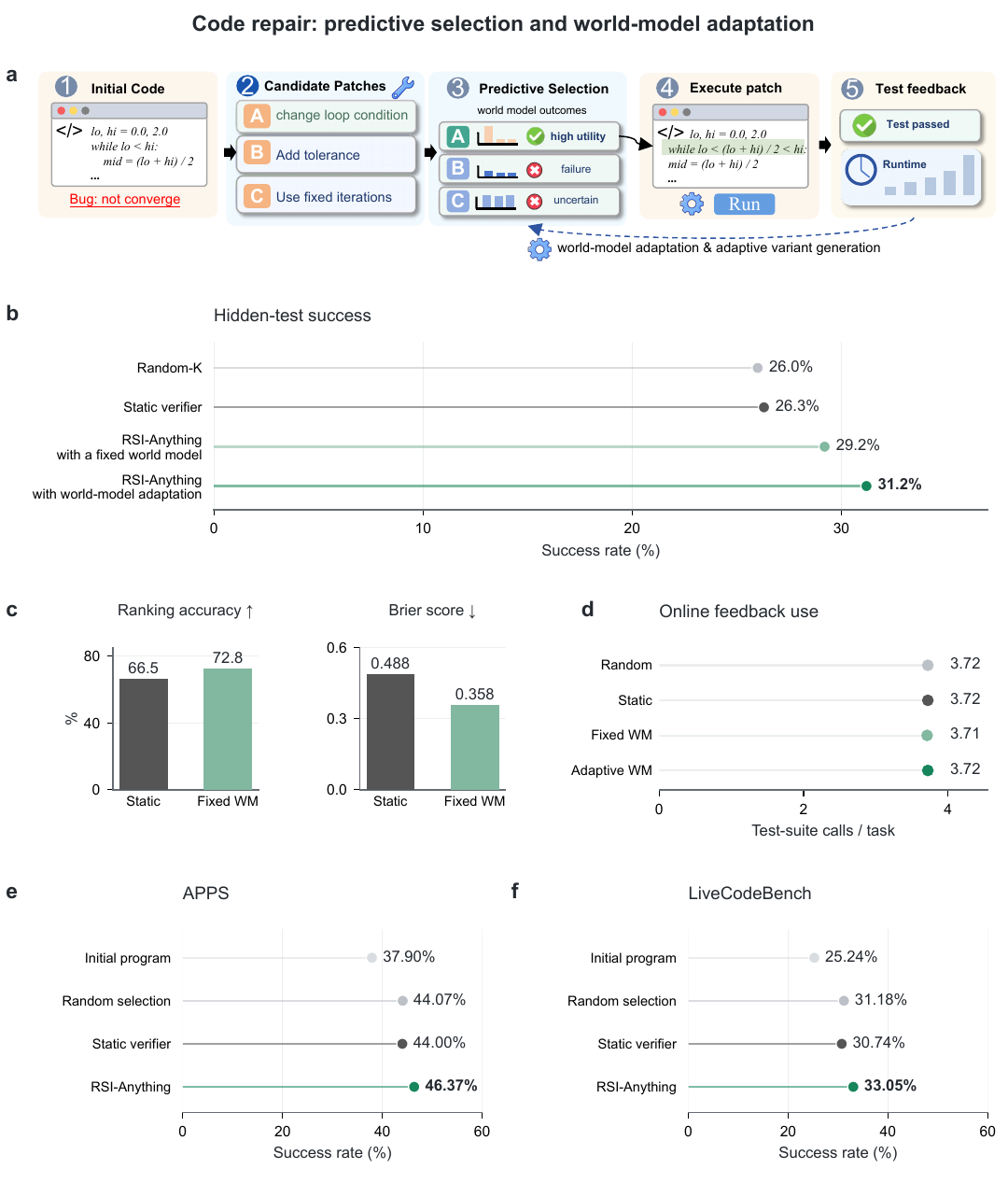}
\caption{Prediction-guided selection and world-model adaptation in code repair. \textbf{a}, Illustrative repair of a non-convergent loop, from candidate patches and predicted outcomes to execution and test feedback. A--C depict example patches; the benchmark uses a fixed coder and four candidates per round. Feedback supports predictor updates in the adaptive configuration. Panels b--d report results on 200 CodeContests tasks across three seeds with up to three feedback rounds. \textbf{b}, Final hidden-test all-pass rate. \textbf{c}, Candidate-ranking accuracy and five-class Brier score for the static verifier and the fixed-world-model configuration. Higher ranking accuracy and lower Brier score indicate better predictive performance. \textbf{d}, Mean online test-suite calls per task. \textbf{e,f}, Final full-test success averaged over three seeds on APPS Introductory (1,000 problems) and LiveCodeBench v6 (1,055 problems), respectively. Fixed WM denotes RSI-Anything with a fixed world model (CURE-WM-Lite); Adaptive WM denotes RSI-Anything with world-model adaptation. Green shades identify world-model-based methods; gray shades identify baselines.}
\label{fig:code-results}
\end{figure}

Introducing world-model-guided selection increased final hidden-test success to $29.2\%$ ($175/600$), compared with $26.3\%$ for a capacity-matched static verifier and $26.0\%$ for random selection (Figure~\ref{fig:code-results}; Table~\ref{tab:code-results}). The paired gain over the static verifier was $+2.83$ percentage points (95\% interval $[+0.17,+5.50]$). In this fixed-world-model versus static-verifier comparison, candidate-ranking accuracy increased from $66.5\%$ to $72.8\%$, and the five-class Brier score \cite{brier1950verification} decreased from $0.488$ to $0.358$. Online test-suite usage was similar: $3.71$ calls per task for the fixed-world-model configuration and $3.72$ for the static verifier. These results connect improved prediction and ranking with higher repair success under the same candidate budget.

Feedback-driven world-model adaptation produced an additional gain over fixed-model selection. With the coder and candidate budget retained, RSI-Anything with world-model adaptation reached $31.2\%$ hidden-test success, exceeding the fixed-world-model configuration by $2.0$ percentage points. Online test-suite usage was $3.72$ calls per task. Execution feedback informed program repair and revised the predictor used for subsequent candidate selection. The two comparisons show successive gains from introducing predictive selection and then adapting the world model, with similar online test usage.

On APPS Introductory \cite{hendrycks2021apps} and LiveCodeBench v6 \cite{jain2024livecodebench}, JEPA-Anything-guided candidate selection achieved final full-test success rates of $46.37\%$ and $33.05\%$, respectively, exceeding random selection and the static verifier (Figure~\ref{fig:code-results}e,f). These comparisons used the same coder, with four repair candidates per round and at most three repair rounds.

\begin{table}[!htb]
\centering
\small
\caption{Three-round repair on CodeContests \cite{li2022alphacode} under a shared coder and candidate budget. Both RSI-Anything configurations use world-model-guided candidate selection: one keeps the world model fixed (CURE-WM-Lite), and the other additionally adapts it using feedback.}
\label{tab:code-results}
\begin{tabular}{>{\raggedright\arraybackslash}p{0.52\linewidth}cc}
\toprule
\textbf{Selection method} & \textbf{Hidden all-pass} & \textbf{Online calls/task} \\
\midrule
Random-K & 26.0\% & 3.72 \\
Capacity-matched static verifier & 26.3\% & 3.72 \\
RSI-Anything with a fixed world model (CURE-WM-Lite) & 29.2\% & 3.71 \\
\textbf{RSI-Anything with world-model adaptation} & \textbf{31.2\%}& 3.72\\
\bottomrule
\end{tabular}
\end{table}

\subsection{RSI-Anything improves task completion in GUI environments}

We evaluated graphical user interface (GUI) task completion on MiniWoB++ \texttt{click-checkboxes}, which requires selecting specified checkboxes and submitting the selection \cite{liu2018workflow}, and on OSWorld \cite{xie2024osworld}. The checkbox task additionally provides action-level diagnostics. RSI-Anything used interaction feedback to update both the policy and the world model, retaining these updates to guide subsequent GUI actions.

\begin{figure}[!htb]
\centering
\includegraphics[width=\linewidth]{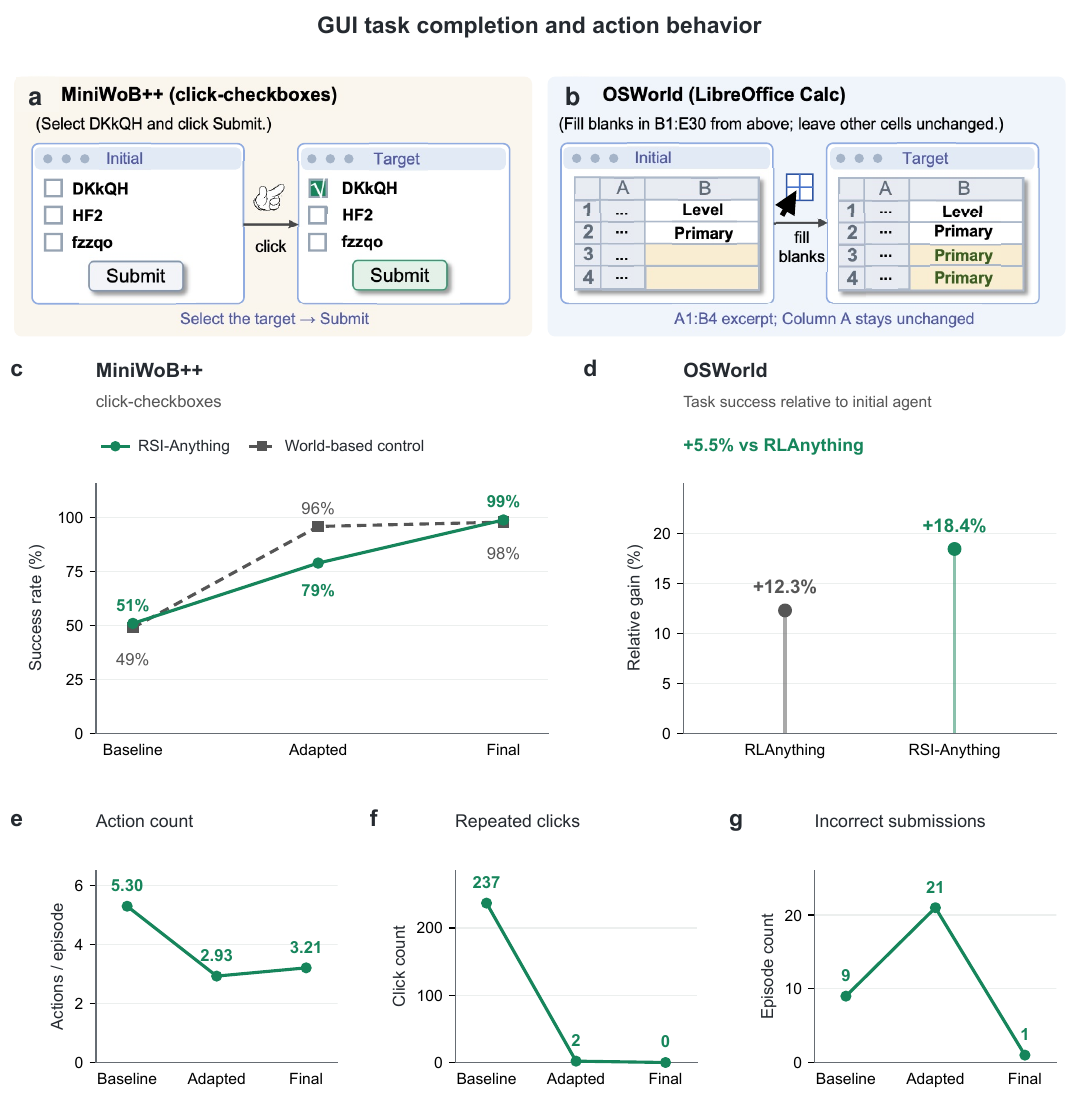}
\caption{Task completion and action behavior in GUI environments. \textbf{a,b}, Illustrative MiniWoB++ checkbox selection and OSWorld spreadsheet completion tasks, showing initial and target states. \textbf{c}, MiniWoB++ \texttt{click-checkboxes} success across baseline, adapted, and final configurations for RSI-Anything and the World-based control. \textbf{d}, OSWorld success-rate gains relative to the initial agent. The annotation gives the additional relative gain of RSI-Anything over RLAnything; all gains are relative percentages. \textbf{e--g}, Mean actions per episode, repeated-click count, and number of episodes with incorrect submissions, respectively, for the RSI-Anything configurations in \textbf{c}. Lines connect evaluated configurations.}
\label{fig:gui-results}
\end{figure}

On \texttt{click-checkboxes}, success increased from $51\%$ at baseline to $79\%$ in the adapted configuration and $99\%$ in the final RSI-Anything configuration (Figure~\ref{fig:gui-results}c). Mean actions per episode decreased from $5.30$ to $3.21$, a $39.4\%$ reduction from baseline; recorded repeated clicks fell from $237$ to $0$, and episodes with incorrect submissions fell from $9$ to $1$ (Figure~\ref{fig:gui-results}e--g). The adapted configuration used fewer actions ($2.93$ per episode) but produced $21$ incorrect-submission episodes. The final configuration therefore combined higher success with fewer submission errors despite a modest increase in action count relative to the adapted configuration. The World-based control achieved $49\%$, $96\%$, and $98\%$ success across the three configurations (Figure~\ref{fig:gui-results}c).

On OSWorld, RSI-Anything improved task success rate by $5.5\%$ relative to RLAnything \cite{wang2026rlanything} and by $18.4\%$ relative to the initial agent (Figure~\ref{fig:gui-results}d). RLAnything improved success by $12.3\%$ relative to the initial agent.

\subsection{Successive policy updates improve laboratory planning}
We next examined whether successive policy updates with RSI-Anything improved physical-domain planning. LabHorizon evaluates laboratory actions subject to reagent, instrument, and ordering constraints through Level-1 visually grounded next-action selection and Level-2 protocol-aligned sequence generation \cite{labplanning2026}. We followed five consecutive versions of a 4B policy from one self-improvement process and evaluated them in development order on the same 200-task F200 Level-2 benchmark. The Final Score combines action-sequence similarity and parameter accuracy.

\begin{figure}[!htb]
\centering
\includegraphics[width=\linewidth]{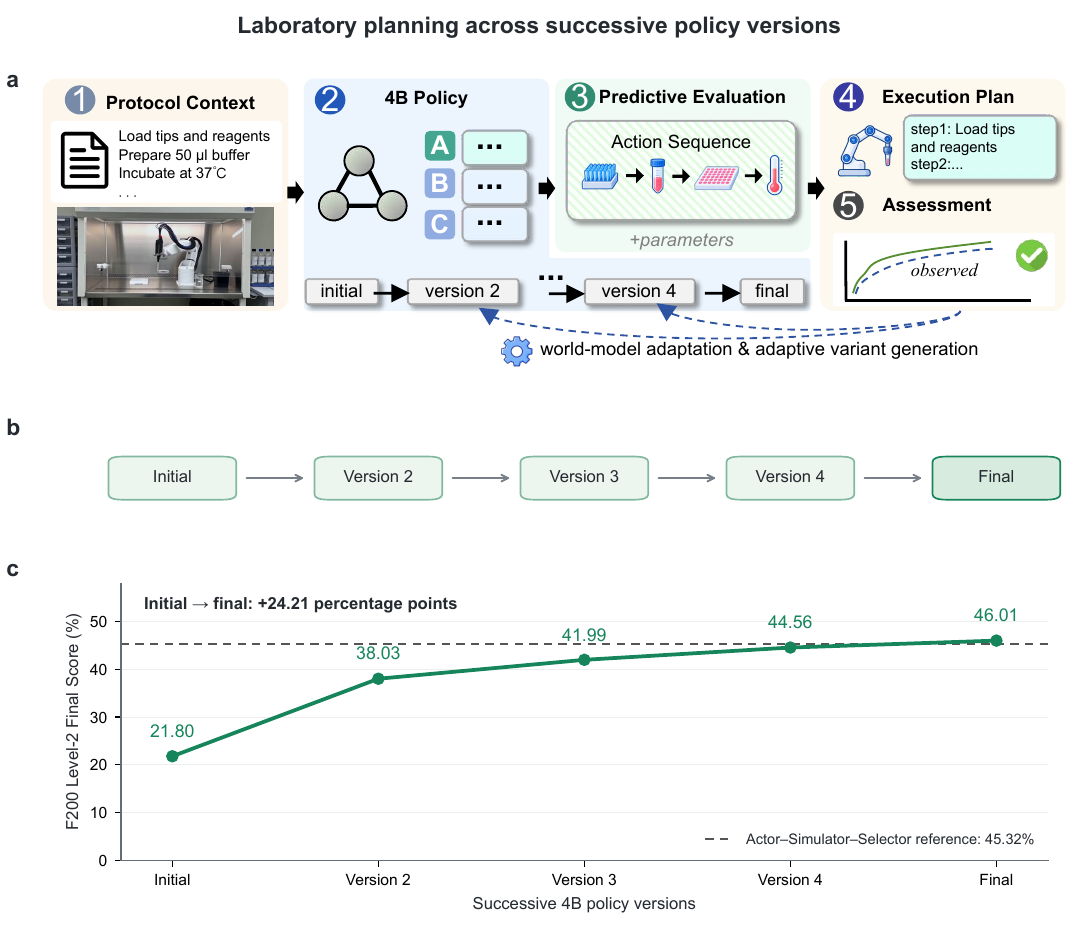}
\caption{Successive policy improvement with RSI-Anything on the 200-task LabHorizon F200 Level-2 benchmark \cite{labplanning2026}. \textbf{a}, Illustrative planning workflow connecting protocol context, a 4B policy, predictive evaluation of action sequences and parameters, an execution plan, and assessment. Dashed feedback paths depict update routes in the framework; the quantitative comparison below follows retained policy versions. \textbf{b}, Five consecutive versions of the same 4B policy from one self-improvement process. \textbf{c}, Final Score for each version, combining action-sequence similarity and parameter accuracy. The dashed line marks the Actor--Simulator--Selector reference at $45.32\%$. Versions are ordered by development sequence; spacing is schematic.}
\label{fig:labhorizon-results}
\end{figure}

The Final Score increased across all five evaluated versions: $21.80\%$, $38.03\%$, $41.99\%$, $44.56\%$, and $46.01\%$ (Figure~\ref{fig:labhorizon-results}). The final version improved on the initial version by $24.21$ percentage points. This progression shows continued improvement in protocol-aligned planning as successive policy versions were retained within the RSI-Anything process.

For comparison with reference systems, the final 4B version achieved a Level-2 score of $46.01\%$, compared with $45.32\%$ for the Actor--Simulator--Selector reference \cite{labplanning2026}, a difference of $0.69$ percentage points. On Level-1 next-action selection, RSI-Anything achieved an accuracy of $0.700$, compared with $0.665$ for the reference (Table~\ref{tab:labhorizon-benchmark}).

\begin{table}[!htb]
\centering
\small
\setlength{\tabcolsep}{4.5pt}
\caption{LabHorizon benchmark scores for reference systems \cite{labplanning2026} and RSI-Anything.}
\label{tab:labhorizon-benchmark}
\begin{tabular}{>{\raggedright\arraybackslash}p{0.62\linewidth}cc}
\toprule
\textbf{System} & \textbf{L1 Acc.} & \textbf{L2 Final} \\
\midrule
Grok 4.3 \cite{xai2026grok43} & 0.555 & 0.3244 \\
Gemini 3.1 Pro \cite{google2026gemini31pro} & 0.465 & 0.3263 \\
GPT-5.5 \cite{openai2026gpt55} & 0.535 & 0.2276 \\
Kimi K2.6 \cite{kimi2026k26} & 0.550 & 0.3150 \\
Qwen3.6-35B-A3B \cite{qwen2026qwen36}, direct & 0.475 & 0.2534 \\
Qwen3.6-35B-A3B, trained & 0.635 & 0.4100 \\
Actor--Simulator--Selector reference & 0.665 & 0.4532 \\
\textbf{RSI-Anything (4B)} & \textbf{0.700} & \textbf{0.4601} \\
\bottomrule
\end{tabular}
\end{table}

\subsection{Prediction- and feedback-guided search improves reaction yields at matched experiment counts}

We examined the use of limited external feedback in experimental science by comparing reaction-condition search at matched experiment counts. RSI-Anything used world-model yield predictions to rank candidate conditions for experimental evaluation. Each measured yield extended the observation history used to guide subsequent condition choices, linking prediction, experiment, and the next selection. We evaluated this process in two Suzuki--Miyaura screens, comparing RSI-Anything with Bayesian optimization \cite{shields2021bayesian} within each screening protocol.

\begin{figure}[!htbp]
\centering
\includegraphics[width=\linewidth,height=0.72\textheight,keepaspectratio]{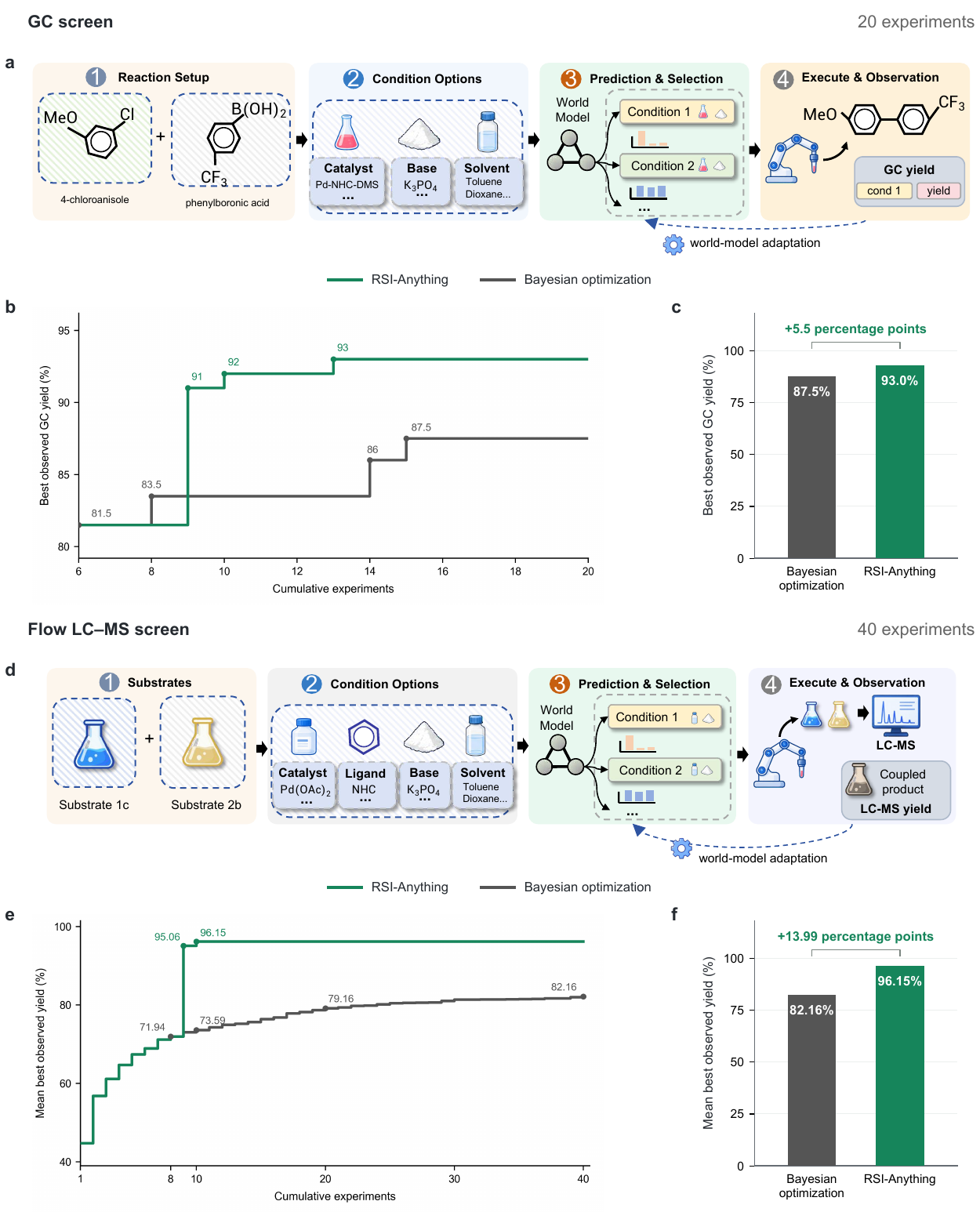}
\caption{Reaction-condition optimization with RSI-Anything (green) and Bayesian optimization (gray). \textbf{a}, Illustrative GC-screen workflow connecting the reaction setup, catalyst/base/solvent options, prediction-guided condition selection, and measured yield. The reaction couples 4-chloroanisole and 4-(trifluoromethyl)phenylboronic acid using a Pd--NHC--DMS precatalyst. \textbf{b}, Best observed GC yield versus cumulative experiment count. \textbf{c}, GC yield after 20 experiments: $93.0\%$ for RSI-Anything and $87.5\%$ for Bayesian optimization. \textbf{d}, Illustrative flow-screen workflow for substrates 1c and 2b, with Pd(OAc)$_2$, ligand, base, and solvent selection followed by LC--MS analysis. \textbf{e}, Mean best observed yield versus cumulative experiment count in the flow LC--MS screen. \textbf{f}, Yield after 40 experiments: $96.15\%$ and $82.16\%$, respectively. Brackets indicate gains of $5.5$ and $13.99$ percentage points within the corresponding screen. Condition examples and feedback arrows illustrate the framework; curves and bars report the two separate screens.}
\label{fig:chem-results}
\end{figure}

Figure~\ref{fig:chem-results} shows yield progression with cumulative experiments (b, e) and endpoint comparisons at matched experiment counts (c, f). After 20 experiments in the GC-analyzed Pd--NHC--DMS screen, the best observed yield reached $93.0\%$ with RSI-Anything and $87.5\%$ with Bayesian optimization, a gain of $5.5$ percentage points. In the flow LC--MS screen using Pd(OAc)$_2$, the mean best observed yield after 40 experiments was $96.15\%$, compared with $82.16\%$ for Bayesian optimization, a gain of $13.99$ percentage points.

At the matched endpoints within each screen, prediction- and feedback-guided search achieved higher best-yield metrics from the same number of experiments. These results support using world-model forecasts together with measured outcomes to guide condition selection under a limited experimental budget.

\subsection{An end-to-end application of RSI-Anything in immune science}
We applied RSI-Anything to an end-to-end immune-science workflow that connects persistent workflow development, predictive world-model construction, and evidence-guided scientific analysis. During construction, Agent Genesis searched scientific-agent workflows and model configurations and selected a model-building agent workflow \cite{evoiwm2026}. The retained workflow produced an action-conditioned immune world model. During downstream analysis, an analysis agent used this model with fixed parameters to coordinate simulation, evaluation, tool use, analysis revision, and hypothesis prioritization.

\begin{figure}[!htb]
\centering
\includegraphics[width=\linewidth]{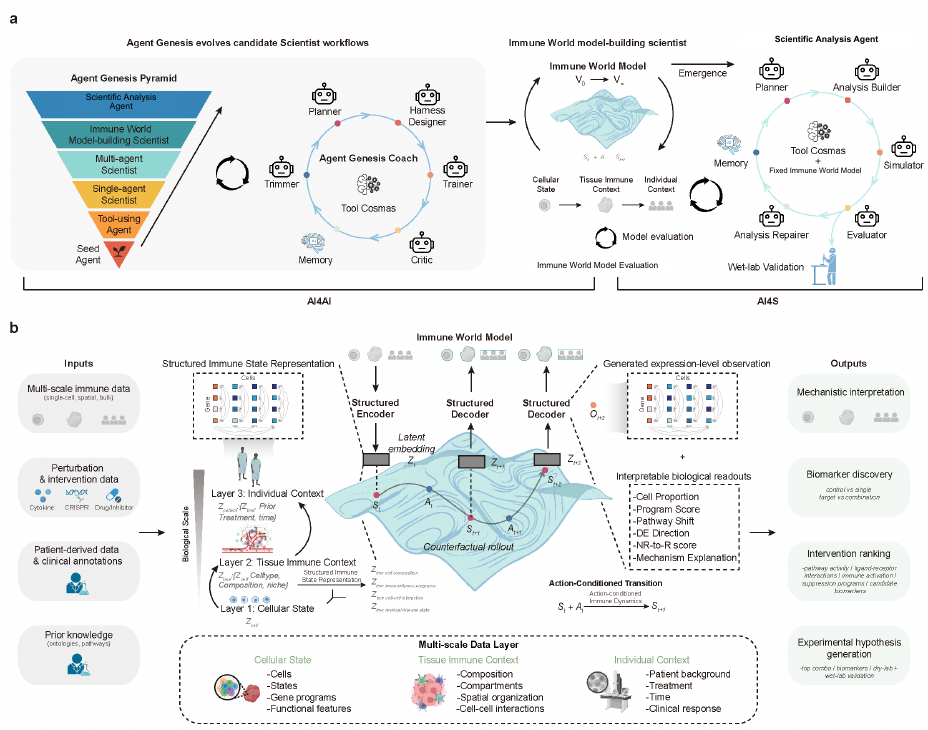}
\caption{An end-to-end application of RSI-Anything in immune science. \textbf{A}, evolution of scientific-agent workflows, construction of the immune world model, and model-guided analysis. \textbf{B}, action-conditioned transitions over structured cellular, tissue, and individual immune states, with readouts for interpretation, ranking, and hypothesis generation. Panels A and B adapted from Figure~1 of \cite{evoiwm2026}.}
\label{fig:immune-application}
\end{figure}

The model represents cellular, tissue, and individual immune states and predicts the consequences of explicit interventions (Figure~\ref{fig:immune-application}). Measured perturbation evidence supports model development, and withheld interventions and biological contexts assess forecasting and action composition. In the analysis stage, model forecasts and experimental evidence inform intervention ranking and self-correction. The construction stage retains the selected model-building workflow and resulting predictor; subsequent analysis uses the constructed predictor to revise candidate interventions and hypotheses.

In therapeutic hypothesis generation, self-correction removed the screened cytokine-pair candidates, and the integrated evidence led to a prospective nomination combining SIRP$\alpha$ inhibition with IL-36$\gamma$-axis modulation \cite{evoiwm2026}. This case connects a retained scientific-agent workflow, a predictive world model, and evidence-guided analysis within one scientific workflow, culminating in a concrete hypothesis for experimental testing.

\FloatBarrier
\Needspace{6\baselineskip}
\section{Related Work}

\subsection{Recursive self-improvement}

Recursive self-improvement (RSI) has evolved from improving reasoning and code to adapting interactive agents and their underlying systems. In coding, CURE co-evolves code generation and unit testing through execution feedback \cite{wang2025cure}, while Self-Improving Coding Agent and Darwin Godel Machine iteratively modify agent implementations and retain empirically validated improvements \cite{robeyns2025sica,zhang2025dgm}. AIDE$^2$ extends this approach to self-improving AI research agents \cite{srikanth2026aide2}.
Beyond code, Voyager accumulates reusable executable skills \cite{wang2023voyager}, while GenEnv and Agent0 adapt task curricula to evolving agent capabilities \cite{guo2025genenv,xia2025agent0}. RLAnything jointly optimizes the policy, reward model, and environment \cite{wang2026rlanything}, and OpenClaw-RL enables online policy improvement from user interactions and tool feedback \cite{wang2026openclawrl}.
Long-horizon RSI further requires improvements to persist across extended interactions. Recuris couples Experiential and Working Memory with validation-gated memory evolution \cite{yu2026recuris}, while ScienceBuddy introduces nested self-improvement through harness refinement and model reinforcement learning \cite{xue2026sciencebuddy}. These approaches improve agent code, policies, memories, harnesses, or training environments. RSI-Anything extends this line by explicitly incorporating predictive world models into the recursive loop: predictions guide candidate selection, observations assess model reliability, and policy--model compatibility determines which updated components are retained for subsequent interactions.

\subsection{Verification and policy self-improvement}
Verifiers, self-training, preference optimization, and verbal feedback turn evaluated trajectories into signals for selection and learning \cite{cobbe2021verifiers,zelikman2022star,rafailov2023dpo,gulcehre2023rest,shinn2023reflexion}. CURE \cite{wang2025cure} co-evolves a coder and unit tester, providing the ReasonFlux-Coder policy used in our code evaluation. RLAnything jointly optimizes the policy, reward model, and environment through feedback \cite{wang2026rlanything}. These approaches motivate the feedback-to-policy part of RSI-Anything. Our framework additionally makes world-model reliability and policy--model compatibility explicit in update acceptance: changing the policy changes the candidate distribution on which its predictor must be evaluated.

\subsection{Predictive world models}
World Models \cite{ha2018worldmodels}, MuZero \cite{schrittwieser2020muzero}, Dreamer \cite{hafner2025dreamer}, and Genie \cite{bruce2024genie} establish predictive environments as a basis for control, planning, and interactive generation. Joint-embedding predictive architectures learn latent targets without reconstructing every observation detail \cite{assran2023ijepa,bardes2024vjepa,assran2025vjepa2}, and JEPA-Anything extends this line through OPF across heterogeneous data types \cite{jepaanything2026}. Building on these predictive capabilities, RSI-Anything organizes how predictions and observed outcomes guide action selection and the acceptance of subsequent policy or world-model versions.

\subsection{Scientific and laboratory agents}
Vision-language-action and cross-embodiment datasets motivate action-conditioned models for physical systems \cite{brohan2023rt2,openx2023,khazatsky2024droid}. LabHorizon measures visual next-action prediction and protocol-aligned planning \cite{labplanning2026}. In chemistry, language-model tools, yield predictors, Bayesian optimization, and robotic laboratories provide complementary components for selecting and testing reactions \cite{boiko2023autonomous,bran2024chemistry,schwaller2021yield,shields2021bayesian,burger2020robotic,merchant2023alab}. ReactionT5 \cite{sagawa2025reactiont5} is a reaction-pretrained transformer that supports product, retrosynthesis, and yield prediction; our chemistry adapter combines this reaction-prediction backbone with the predictive design of JEPA-Anything \cite{jepaanything2026} for action-conditioned yield forecasting. Single-cell foundation models and perturbation predictors support cellular-response forecasting \cite{cui2024scgpt,theodoris2023geneformer,roohani2024gears,bunne2023cellot}. RSI-Anything connects these domain capabilities through a common transition and feedback contract, so that the outcome of a program, experiment, or intervention can inform candidate selection and subsequent system updates.

\section{Discussion}

The results support two roles for world models in recursive improvement: informing which actions receive external feedback and providing a predictor that can itself improve from that feedback. Predictive selection and model adaptation yielded successive gains in code repair, while retained policy versions improved laboratory planning. In GUI tasks, RSI-Anything updated both the policy and the world model and achieved higher task completion, with fewer repeated clicks and incorrect-submission episodes on the checkbox task. Together with measured-feedback search and the end-to-end immune-science workflow, these findings connect prediction-guided decisions to learning and retained system changes. The common contribution is a procedure for using prediction--observation pairs to guide current actions and assess changes to the system that will make subsequent decisions.

The design links decisions about actions to decisions about system versions. Pairing predictions with observations allows the same experience to inform candidate utility and predictive reliability. Policies and world models can change on different schedules: a reliable predictor can support policy or workflow development, and new outcome evidence can motivate predictor adaptation while the policy remains fixed. Reliability checks govern when model adaptation is attempted, and component and compatibility checks govern which version is carried forward. Retaining accepted versions makes subsequent experience depend on the system's accumulated improvements.

Domain generality concerns the reuse of this improvement mechanism across domain-specific world models, policies, and evaluation procedures. The evaluations support predictive selection, predictor adaptation, successive policy improvement, and feedback-guided search in their respective settings. Joint multiround evaluation is needed to establish sustained mutual improvement of the policy and world model within a single task. The present laboratory results assess protocol-aligned planning; reliable physical execution requires evaluation on the relevant equipment and workflows. The chemistry results cover two Suzuki--Miyaura screens under the reported experiment budgets, and the therapeutic hypothesis generated in the immune-science application awaits experimental testing.

\section{Experimental Methods}

The domain settings instantiate the shared framework with different states, action spaces, prediction targets, and feedback sources. Our code execution predictor, W1, supplies execution-class probabilities for candidate--test pairs, ReactionT5--JEPA-Anything supplies a yield distribution, and the immune world model supplies immune-state and biological readouts. Table~\ref{tab:implementation} summarizes the domain-specific models, prediction targets, feedback sources, and update pathways evaluated in each application. The following sections describe the evaluation settings and statistical reporting.

The experiments instantiate distinct parts of the full protocol. Code repair holds the coder fixed while testing predictive selection and predictor adaptation; laboratory planning follows five successive policy versions. GUI evaluations measure task completion and action behavior as RSI-Anything updates both the policy and the world model. Chemistry tests feedback-guided condition search, and the immune-science application separates workflow evolution and model construction from downstream analysis with a fixed predictor.

\begin{table}[!htb]
\centering
\footnotesize
\setlength{\tabcolsep}{3pt}
\caption{Domain bindings and reported update scope. The world model predicts consequences; the policy proposes or selects actions.}
\label{tab:implementation}
\begin{tabular}{>{\raggedright\arraybackslash}p{0.11\linewidth}
                >{\raggedright\arraybackslash}p{0.23\linewidth}
                >{\raggedright\arraybackslash}p{0.22\linewidth}
                >{\raggedright\arraybackslash}p{0.18\linewidth}
                >{\raggedright\arraybackslash}p{0.18\linewidth}}
\toprule
\textbf{Setting} & \textbf{State; action} & \textbf{World model; prediction} & \textbf{Policy / workflow; feedback-driven changes} & \textbf{Feedback; evaluation} \\
\midrule
Code repair & Program, tests, and feedback; candidate revision & JEPA-Anything-based W1 adapted from Qwen2.5-Coder-3B-Instruct \cite{hui2024qwen25coder} with LoRA \cite{hu2022lora}; five-class probabilities & Fixed ReasonFlux-Coder-7B; fixed or feedback-adapted W1 & Execution-test feedback; final test-suite success and online test usage \\
GUI interaction & Interface state and task; GUI action & GUI outcome predictor; consequences of candidate interface actions & Action policy and world model updated through interaction feedback & Task outcomes; completion and action-level diagnostics \\
Lab planning & Protocol context; candidate action sequence & JEPA-Anything-based protocol model; predicted consequences of a candidate sequence & 4B planning policy; five consecutive versions in one improvement process & Development feedback; L1 next-action accuracy and F200 L2 Final Score \\
Chemistry & Reaction; condition vector & ReactionT5--JEPA-Anything; yield distribution & Measured yields guide subsequent condition choices & Measured GC/LC--MS yields; best observed yield at matched experiment counts \\
Immune science & Cellular, tissue, and individual context; intervention & Immune world model \cite{evoiwm2026}; multiscale transition and biological readouts & Workflow evolution and world-model construction; fixed predictor during downstream analysis & Measured perturbations support development; withheld contexts assess forecasting \\
\bottomrule
\end{tabular}
\end{table}

\subsection{Domain-specific evaluation settings}

\paragraph{Code repair.}
The state contains the current program, generated tests, and observed feedback; each action is a candidate revision. The ReasonFlux-Coder-7B policy proposes four revisions and is held fixed \cite{wang2025cure}. W1 is our 3B execution-outcome world model, built using the predictive design of JEPA-Anything \cite{jepaanything2026} and adapted from Qwen2.5-Coder-3B-Instruct \cite{hui2024qwen25coder} with LoRA \cite{hu2022lora}. For every candidate--test pair, it returns the probability of each execution-result class. Both RSI-Anything configurations aggregate these probabilities to rank the four revisions before one revision is executed. The fixed-world-model configuration (CURE-WM-Lite) retains W1, whereas the world-model-adaptation configuration updates W1 with the coder held fixed. We evaluated 200 CodeContests tasks \cite{li2022alphacode} across three seeds and at most three feedback rounds. The comparison included both RSI-Anything configurations, random selection, and a capacity-matched static verifier. Metrics were final hidden-test all-pass rate, online test calls per task, candidate-ranking accuracy, and five-class Brier score. Online tests supply feedback during selection; hidden tests supply the final reported success measure.

We additionally evaluated APPS Introductory (1,000 problems) and LiveCodeBench v6 (1,055 problems) using seeds 0, 1, and 2. Repair methods shared the initial program for each problem and seed and used the same ReasonFlux-Coder-7B policy, with four repair candidates per round and at most three repair rounds. Public-test feedback guided repair, which stopped when all public tests passed. Final success required passing the complete official test suite.

\paragraph{GUI agents.}

We evaluated the MiniWoB++ \texttt{click-checkboxes} task \cite{liu2018workflow} and OSWorld \cite{xie2024osworld}. World-model predictions guide GUI action selection, and observed interaction outcomes provide feedback for updating both the action policy and the world model. The retained policy--model pair is used in subsequent interactions. The checkbox evaluation compares baseline, adapted, and final configurations using task success rate, mean actions per episode, repeated-click count, and the number of episodes with incorrect submissions. Success rates are also shown for the World-based control. The action and error diagnostics in Figure~\ref{fig:gui-results}e--g describe the RSI-Anything configurations. The OSWorld comparison includes the initial agent, RLAnything \cite{wang2026rlanything}, and RSI-Anything. For success rates $s_{\mathrm{method}}$ and $s_{\mathrm{reference}}$, relative improvement is $100(s_{\mathrm{method}}/s_{\mathrm{reference}}-1)\%$. Figure~\ref{fig:gui-results}d uses the initial agent as the common reference; the additional RSI-Anything comparison uses RLAnything as its reference.

\paragraph{Laboratory planning.}
LabHorizon evaluates Level-1 next-action accuracy and Level-2 protocol-aligned action sequences on its 200-task F200 setting \cite{labplanning2026}. Level-2 Final Score combines action-sequence similarity and parameter accuracy. The RSI-Anything policy versions use a 4B model to generate plans and a JEPA-Anything-based protocol world model to predict their consequences. The F200 evaluation follows five consecutive 4B policy versions from the same RSI-Anything self-improvement process, ranging from $0.2180$ to $0.4601$. The Actor--Simulator--Selector reference uses a trained Qwen3.6-35B-A3B Actor \cite{qwen2026qwen36} and Gemini 3.1 Pro as Simulator and Selector \cite{google2026gemini31pro}, with a reported LabHorizon Level-2 score of $0.4532$ \cite{labplanning2026}. Level-1 next-action accuracy was $0.700$ for RSI-Anything and $0.665$ for the reference.

\paragraph{Reaction optimization.}
We construct the ReactionT5--JEPA-Anything adapter by combining the ReactionT5 reaction-prediction backbone \cite{sagawa2025reactiont5} with the predictive design of JEPA-Anything \cite{jepaanything2026}. It maps a serialized reaction $r$ and candidate condition vector $c$ to a predicted yield distribution $p_{\phi}(y\mid r,c)$. One screen varied Pd--NHC--DMS precatalyst, base, and solvent for the 4-chloroanisole coupling and measured GC analytical yield. A second screen used Pd(OAc)$_2$ with ligand, base, and solvent choices in flow, measured by LC--MS. Bayesian optimization is the baseline in both screens. Results are reported as best observed or mean best observed yield for each separate protocol. Measured yields guide subsequent condition choices, and performance is summarized by the best observed yield at matched experiment counts.

\paragraph{End-to-end immune-science application.}
The immune-science application uses an action-conditioned immune world model that encodes cellular state, tissue immune context, and individual context, and treats cytokines, genetic perturbations, or candidate therapies as explicit actions \cite{evoiwm2026}. Agent Genesis evolved scientific-agent workflows and model configurations, producing a model for withheld-intervention forecasting, cross-scale prediction, and action composition. The selected model-building agent workflow is the retained policy configuration in this construction stage. During downstream intervention analysis, the resulting world-model parameters are frozen while the analysis state, candidate interventions, and hypothesis ranking evolve through model inference and self-correction. Therapeutic-hypothesis generation combined measured perturbation evidence with this model-guided analysis. The resulting prospective nomination combines SIRP$\alpha$ inhibition with IL-36$\gamma$-axis modulation for subsequent experimental testing.

\subsection{Statistical reporting}
Success rates are reported over 600 CodeContests task-seed evaluations. We report the paired percentage-point difference between RSI-Anything with a fixed world model and the static verifier, together with its 95\% interval. Tasks recur across seeds, so task-level resampling is the unit for clustered inference. APPS and LiveCodeBench report the arithmetic mean of final full-test success rates across three seeds, corresponding to 3,000 and 3,165 problem--seed evaluations, respectively. LabHorizon F200 reports mean scores over 200 tasks \cite{labplanning2026}. The $24.21$- and $0.69$-percentage-point gains are differences between these benchmark means. Chemistry reports the observed running best within each optimization sequence, with Bayesian optimization as the comparator. Evaluation settings for the immune-science application are described in \cite{evoiwm2026}.

\begingroup
\setlength{\bibsep}{2pt}
\bibliography{references}
\endgroup

\clearpage
\appendix
\section{Author Affiliations and Correspondence}
\label{app:affiliations}

\begin{enumerate}[label=\textsuperscript{\arabic*},leftmargin=1.6em,itemsep=0.35em]
    \item PhAI Labs
    \item The Chinese University of Hong Kong
    \item University of Macau
    \item Shanghai Jiao Tong University
    \item Fudan University
    \item University of Oxford
    \item Stanford University
    \item Peking University
\end{enumerate}

\subsection*{Correspondence}

\noindent Correspondence:
\href{mailto:pheng@cse.cuhk.edu.hk}{\texttt{pheng@cse.cuhk.edu.hk}} (P.A.H.);
\href{mailto:yin@phai-labs.com}{\texttt{yin@phai-labs.com}} (Z.Y.);
\href{mailto:wuyc@phai-labs.com}{\texttt{wuyc@phai-labs.com}} (Y.W.); and
\href{mailto:yang@phai-labs.com}{\texttt{yang@phai-labs.com}} (L.Y.).

\end{document}